\documentclass[letter, 10 pt, conference]{ieeeconf}  

\IEEEoverridecommandlockouts                              

\usepackage{graphics} 
\usepackage{epsfig} 
\usepackage{amsmath} 
\usepackage{amssymb}  
\usepackage{bm}
\usepackage{siunitx}
\usepackage{enumerate}
\usepackage{multirow}
\usepackage{tikz}
\usepackage{pgfplots}
\usepackage{algorithm}
\usepackage[noend]{algpseudocode}
\usepackage{hyperref}

\newcommand\copyrighttext{\footnotesize \textcopyright 2022 IEEE. Personal use of this material is permitted. Permission from IEEE must be obtained for all other uses, in any current or future media, including reprinting/republishing this material for advertising or promotional purposes, creating new collective works, for resale or redistribution to servers or lists, or reuse of any copyrighted component of this work in other works.
DOI:\href{tba}{tba}}
\newcommand\copyrightnotice{%
\begin{tikzpicture}[remember picture,overlay]
\node[anchor=south,yshift=10pt] at (current page.south) {\fbox{\parbox{\dimexpr\textwidth-\fboxsep-\fboxrule\relax}{\copyrighttext}}};
\end{tikzpicture}
}

\title{\LARGE \bf
ExAgt: Expert-guided Augmentation for Representation Learning of Traffic Scenarios   
}

\author{Lakshman Balasubramanian$^{*1}$, Jonas Wurst$^{*1}$, Robin Egolf$^{1}$,  Michael Botsch$^{1}$, \\ Wolfgang Utschick$^{2}$ and Ke Deng$^{3}$
\thanks{*Equal contribution}
\thanks{$^{1}$CARISSMA Institute of Automated Driving, Technische Hochschule Ingolstadt, 85049 Ingolstadt, Germany {\tt\small {firstname.lastname}@thi.de}}
\thanks{$^{2}$Technical University of Munich, 80333 Munich, Germany {\tt\small {utschick}@tum.de}}
\thanks{$^{3}$ Royal Melbourne Institute of Technology {\tt\small {ke.deng}@rmit.edu.au}}
}

\begin{document}
\bstctlcite{IEEEexample:BSTcontrol} 
\DeclareRobustCommand {\expag}{\raisebox{2pt}{\tikz{\draw[red,solid,line width=0.9pt](0,0) -- (5mm,0);}}}
\DeclareRobustCommand {\baseag}{\raisebox{2pt}{\tikz{\draw[blue,solid,line width=0.9pt](0,0) -- (5mm,0);}}}

\maketitle
\copyrightnotice

\thispagestyle{empty}
\pagestyle{empty}

\begin{abstract}

Representation learning in recent years has been addressed with self-supervised learning methods. The input data is augmented into two distorted views and an encoder learns the representations that are invariant to distortions -- cross-view prediction. Augmentation is one of the key components in cross-view self-supervised learning frameworks to learn visual representations. This paper presents ExAgt, a novel method to include expert knowledge for augmenting traffic scenarios, to improve the learnt representations without any human annotation. The expert-guided augmentations are generated in an automated fashion based on the infrastructure, the interactions between the EGO and the traffic participants and an ideal sensor model. The ExAgt method is applied in two state-of-the-art cross-view prediction methods and the representations learnt are tested in downstream tasks like classification and clustering. Results show that the ExAgt method improves representation learning compared to using only standard augmentations and it provides a better representation space stability. The code is available at \url{https://github.com/lab176344/ExAgt}.

\end{abstract}

\section{INTRODUCTION}

Representation learning is seen as the key factor for the success of deep learning-based methods. In autonomous driving and vehicle safety applications deep learning methods are used in perception~\cite{Fayyad2020,Arnold2019}, motion planning~\cite{Liu2021}, traffic scenario classification~\cite{Beglerovic2019} and clustering~\cite{Zhao2021,Balasubramanian2021}. This work focuses on representation learning for traffic scenarios. Traffic scenario  learning methods are important for components like motion forecasting, scenario-based validation, etc. The State-Of-The-Art (SOTA) traffic scenario learning methods use deep learning-based architectures, with large labelled datasets if the task to learn is classification or use engineered loss functions with custom architectures if the task to solve is clustering and outlier detection. This procedure is different to domains like Computer Vision (CV) and Natural Language Processing (NLP), where pre-training~\cite{Zoph2020,Ramanathan2021,Devlin2018,Radford2019} to learn representations is seen as an essential component for solving downstream tasks.

Most of the SOTA CV models use deep learning models pre-trained on Imagenet~\cite{Deng2009} to initialise the networks before fine-tuning on the intended task. A similar trend can be seen in NLP, where Word2Vec~\cite{Mikolov2013}, BERT~\cite{Devlin2018} and GPT~\cite{Radford2019} models are used to initialise task-specific models. In recent years, self-supervised learning methods~\cite{Liu2021a,Hendrycks2019} are proving to be effective pre-training methods to learn representations without any human annotation.
\begin{figure}
	\centering
	\includegraphics[width=0.45\textwidth]{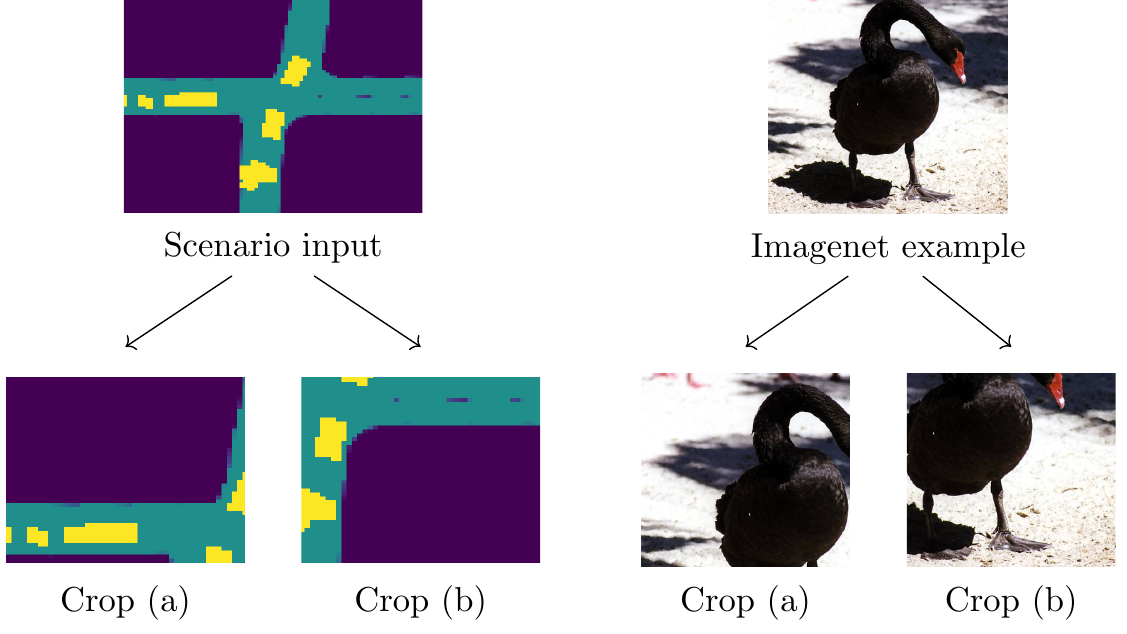}
	\caption{Random crop augmentation in traffic scenarios vs Imagenet images: Imagenet examples has one object of interest, black swan in this example. Hence, making a random crop is meaningful in this case as the object of interest is present in both crops, but in traffic scenarios, several objects relate to each other and induce influence on each other. So, using an augmentation like random crop will result in views which has no connection with the input.}
	\label{fig:baseAugmentation}
    \vspace{-5mm}
\end{figure}

In CV, the most popular self-supervised learning methods use cross-view prediction~\cite{Grill2020,Zbontar2021,Caron2020}. The training of cross-view prediction methods is as follows: the input image is distorted or augmented into two different views. These two views are processed with a shared network to generate two representation vectors, one for each view. The network solves the objective of making these representations similar as they are generated from the same input. There are two important components in the cross-view prediction framework: (a) a learning objective that makes the representation invariant of distortion, (b)  augmentations that are applied to create the two distorted views. Component (a) is domain-independent. But the augmentations to be applied (b) are domain-specific.

This work focuses on learning representations for traffic scenarios, particularly introducing domain-specific augmentations. The traffic scenarios in this work are represented as a sequence of occupancy grids. The input is domain-specific and is different from datasets like Imagenet, CIFAR-100~\cite{Krizhevsky2009}, etc., which are used for self-supervised learning methods. The images from Imagenet and CIFAR have central objects of interest e.\,g., airplanes, dogs. Applying standard augmentation like random crop, gray-scale, etc., \cite{Grill2020} is meaningful, as the distorted views contain a part of the object of interest. This is not necessarily the case in traffic scenarios where there are multiple objects of interest like traffic participants, infrastructure, and spatio-temporal relations (c.\,f. Fig. \ref{fig:baseAugmentation}). Therefore, in ExAgt, standard CV augmentations are extended with expert-guided augmentations.

ExAgt tailors augmentations for learning traffic scenario representations. Two augmentations are proposed for this purpose. The first augmentation is based on the connectivity of the underlying infrastructure and the interaction of traffic participants with the EGO vehicle in a traffic scenario.

This is termed as \textit{connectivity-based augmentation} in this work. The second augmentation introduced in this work is based on simulating sensors in the EGO vehicle with restricted Fields of View (FoV) and ranges. The objects outside the Visible Region~(VR) are removed to create a distorted view. This is termed as \textit{sensor-based augmentation}. ExAgt is designed, such that the goal of traffic scenario learning is aided by the novel augmentations.

The proposed method ExAgt shows superior performance in zero-shot clustering, low-shot supervised learning, and supervised learning, when compared to standard CV augmentations. Pre-trained models as the ones introduced in this work can be used for transfer learning in downstream tasks like scenario clustering and scenario classification.

The main contributions of the work are as follows:
\begin{enumerate}
    \item Novel approach to include expert knowledge for augmenting traffic scenarios.
    \item Representation learning of traffic scenarios using a cross-view prediction framework.
    \item Analysis of the representation space for downstream tasks like clustering and classification.
    \item Investigation of the representation space stability.
\end{enumerate}

The remainder of the paper is organised as follows: Section~\ref{Sec:RW} presents the related work. The proposed method is discussed in Section~\ref{Sec:Met}. Section~\ref{Sec:Exp} illustrates the experiments and analysis on Argoverse dataset \cite{Chang2019}. Finally, the paper is concluded in Section~\ref{Sec:Con}.

\section{RELATED WORKS}
Works focusing on representation learning for traffic scenarios and literatures introducing custom augmentation for self-supervised learning are discussed in this section.
\label{Sec:RW}
\subsubsection{Representation Learning for Traffic Scenarios}
 A multi-step approach for representation learning in traffic scenarios is presented in~\cite{Zhao2021}. As a first step, the sequence of occupancy grids is compressed into a sequence of frame vectors using reconstruction loss and triplet loss. As a second step, the sequence of frame vectors is processed to get a sequence compression vector which is the representation of a scenario. 
 
 In~\cite{Wurst2021}, a triplet learning framework is used for learning representations of road infrastructure. The infrastructure graph and triplet loss are used for representation learning.
 
 The authors in~\cite{Balasubramanian2021} assume a few labelled and unlabelled datapoints are available. Representation for the unlabelled datapoints is learned using self-supervised pre-training and fine-tuning with a Random Forest similarity measure. 
 
 In~\cite{Harmening2020}, a spatio-temporal autoencoder is used to encode the sequence of multi-channel occupancy grids. The latent space of the encoder is the compressed representation of the traffic scenarios and is used for clustering.
 
 All the above-mentioned methods focus on generating representations for traffic scenarios and use them for clustering or outlier detection. In contrast to the methods~\cite{Zhao2021} and~\cite{Balasubramanian2021} the representation learning in this work is a single-step process. Furthermore, this work utilises expert knowledge about the infrastructure and dynamics of the traffic participants. The method proposed in~\cite{Wurst2021} focuses only on the infrastructure part of the traffic scenario. In~\cite{Harmening2020}, an autoencoder with reconstruction loss is used to learn representation. 
 
 \subsubsection{Custom Augmentation in Self-Supervised Networks}
 
 To the best of authors' knowledge, this is the first work to define custom augmentations for traffic scenarios. Custom augmentations are successfully used in domains like NLP and video representation learning.

 \paragraph{NLP Augmentation}
A contrastive self-supervised encoder-only transformer is introduced in~\cite{Fang2020}. The method uses back-translation as the augmentation for text data. The idea is that the back-translated sentence and the original sentence contextually have the same meaning, so these two should be close together in the representation space.

BERT~\cite{Devlin2018} is a bi-directional encoder-only transformer. The augmentation strategy in BERT is randomly masking words in a sentence. BERT aims to predict the masked work.

A similar augmentation strategy is used in GPT~\cite{Radford2019}, where parts of the sentences are masked. The model generates the masked part of the sentence.

 \paragraph{Video Augmentation}
 In~\cite{Recasens2021}, two views of a given video are created by having a narrow temporal slice and a broad temporal slice of the video. The objective is to generalise from the narrow view to the broad view. The temporal structure is used to create a custom augmentation. 
 
 Spatial and temporal augmentation is applied to video clips in~\cite{Qian2021}. Temporally consistent spatial augmentation like colour jitter and gray-scale, are applied across all frames. In the temporal dimension, a fixed number of frames is sampled randomly and the sampling procedure is used to create two views from the same video clip.

\begin{figure*}[htbp]
    \vspace{2mm}
    \centering
    \includegraphics[width=\linewidth]{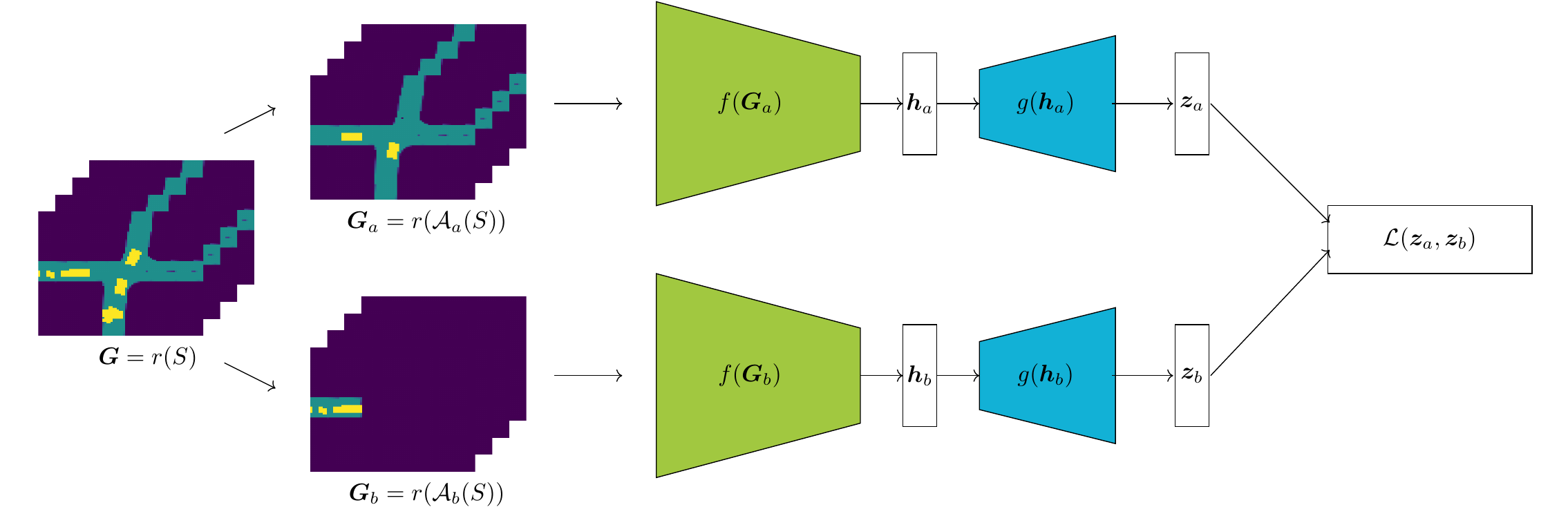}
    \caption{Cross-view prediction framework}
    \label{fig:crossview}
    \vspace{-5mm}
\end{figure*}
 In~\cite{Han2020}, the authors combine different views from optical flow and RGB space of the same clip for a co-training scheme in self-supervised learning, exploiting the complementary information from different views. 
  
The above-mentioned works introduce augmentations based of domain knowledge on the input modality i.\,e., text, video, and speech. Hence, domain-specific augmentations are an important component of self-supervised learning frameworks which is the main focus of this work.

\section{METHODOLOGY}
\label{Sec:Met}
This section discusses the expert-guided augmentation generation for traffic scenarios. Since these augmentations will be used in self-supervised methods, background about self-supervised methods used in this work is also presented.  

\subsection{Preliminaries}
A dataset $\mathcal{D} = \left\{(S_{1}),\ldots,(S_{M})\right\}$ is available, where $S=(\mathcal{O},\mathcal{M})$ is a scenario. The scenario contains an object list $\mathcal{O}=\lbrace o_1,\dots,o_I\rbrace$ and a map $\mathcal{M}=\lbrace m_1,\dots,m_J\rbrace$. Each object $o=(\mathcal{T},\bm{s},c)$ contains a trajectory $\mathcal{T}$, and information about the object size $\bm{s}$ and type of the object $c$. One map element $m=(\mathcal{P},\mathcal{N},I)$ represents a lane piece. It contains the underling polygon $\mathcal{P}$ and connectivity information: the set of neighbouring lane pieces $\mathcal{N}$ and the intersection $I$ which the lane piece $m$ is part of.
Per scenario, the corresponding sequence of occupancy grids $\bm{G} \in\mathbb{R}^{C\times H\times W}$ can be generated by $\bm{G}=r(S)$, with $H\times W$ being the size of the grids over $C$ timestamps.

Let $f(.)$ be a trainable network, realising the mapping from the occupancy grid sequence to the representation vector $\bm{h} = f(\bm{G})$ with $\bm{h} \in \mathbb{R}^{d_\mathrm{r}}$. Like in~\cite{Zbontar2021}, the representation vector $\bm{h}$ is up-projected with the trainable network $g(.)$, leading to the embedding vector $\bm{z} = g(\bm{h})$ with $\bm{z} \in \mathbb{R}^{d_\mathrm{p}}$. 

In self-supervised learning, the goal is to train the networks $f(.)$ and $g(.)$ to generate meaningful representations without the need for any labelled data. Most of these methods use the cross-view prediction framework~\cite{Grill2020,Zbontar2021,Caron2020}. The schematics of such a framework is shown in Fig.~\ref{fig:crossview}. An input is distorted using transformations to generate two augmented views using the transformations $\mathcal{A}_a$ and $\mathcal{A}_b$. The encoder network $f$ uses the two distorted views to produce representations $\bm{h}_a$ and $\bm{h}_b$. The representations $\bm{h}_a$ and $\bm{h}_b$ are up-projected with a projection network $g(.)$ to get the embeddings $\bm{z}_a = g(\bm{h}_a) $ and $\bm{z}_b = g(\bm{h}_b) $ respectively. A learning objective $\mathcal{L}(\bm{z}_a,\bm{z}_b)$ is defined using the embedding.

The objective with the cross-view prediction framework is to learn representations that are invariant to distortions applied to the inputs. This constraint is achieved by making the representation of the distorted view of an input similar to another distorted view of the same image. An undesired outcome of this constrain is that all the representations becoming constant. This is referred to as the collapse of representation in literatures. There are different mechanisms to prevent this collapse by applying constraints in the learning objective $\mathcal{L}$. The following section describes two of the most recent mechanisms.

\subsubsection{Barlow Twins~\cite{Zbontar2021}}

The mechanism used in Barlow Twins is redundancy reduction between the representations of the two distorted views. This is achieved by measuring the cross-correlation matrix between the outputs of $g(\bm{h}_a)$ and $g(\bm{h}_b)$, and making it as close to the identity matrix as possible.

\subsubsection{VICReg~\cite{Bardes2022}}
VICReg uses Variance-Invariance-Covariance Regularisation in the learning objective to prevent collapse. Two regularisations are applied to the representations of the two distorted views. The first maintains the variance of each representation above a certain threshold and the second decorrelates the representation dimensions. 

\subsection{Expert Knowledge-based Augmentations}
Self-supervised learning methods, which are based on cross-view predictions depend on two important components: (a) the learning objective to prevent collapse (b) possible augmentations to the input. Standard augmentations for the cross-view prediction frameworks in CV are random crop, colour jitter, gray-scale, Gaussian blur, mix-up, etc~\cite{Grill2020,Zbontar2021,Caron2020}. But these augmentations cannot be directly used for domain-specific inputs like traffic scenarios. Applying these techniques to traffic scenarios may not aid the learning as exemplified in Fig.~\ref{fig:baseAugmentation}. It shows a single occupancy grid from the sequence of occupancy grids representing a scenario and an image from Imagenet. When applying the random crop augmentation on both of these images, the black swan is present in both the images. However, in the occupancy grid, a random crop creates two views, where the connection between both is lost and can lead to a different interpretation of the scene. This is because the CV datasets like Imagenet, and CIFAR-100, have a central object of interest. For domains like traffic scenarios, there are multiple objects of interest like the infrastructure, traffic participants and the spatio-temporal relations between them. So, only standard CV augmentations might not be sufficient for domain-specific inputs.

In order to create meaningful augmentations, ExAgt introduces two types of domain-specific augmentations. The first augmentation $\mathcal{A}_\mathrm{con}$ is based on the connectivity of the underlying infrastructure and traffic participants interaction with the EGO vehicle. The second augmentation $\mathcal{A}_\mathrm{VR}$ is based on sensor models which restrict the FoV and range of the EGO vehicle perception. 

The standard CV augmentations manipulate only the image $\bm{G}$ as $\tilde{\bm{G}}=\mathcal{A}(r(S))$, without any additional knowledge. In contrast, ExAgt is leveraging expert information, such that the augmentations help in achieving better performance. Therefore, in ExAgt an augmented occupancy grid is generated by $\tilde{\bm{G}}=r(\mathcal{A}(S))$, where $\mathcal{A}$ can be $\mathcal{A}_\mathrm{con}$, $\mathcal{A}_\mathrm{VR}$ or $\mathcal{A}_\cap$, which is the combination of  $\mathcal{A}_\mathrm{con}$ and $\mathcal{A}_\mathrm{VR}$. With the augmented scenarios $S_\mathrm{con} = \mathcal{A}_\mathrm{con}(S)$ and $S_\mathrm{VR} = \mathcal{A}_\mathrm{VR}(S)$ resulting from the augmentations applied to the scenario $S$, the combined augmentation $\mathcal{A}_\cap$ can be defined as 
\begin{equation}
    S_\cap = (\mathcal{O}_\mathrm{con} \cap \mathcal{O}_\mathrm{VR},\mathcal{M}_\mathrm{con} \cap \mathcal{M}_\mathrm{VR}).
\end{equation}

Both augmentations follow the intuition, that certain information of the scenario can be dropped while maintaining some relevant information of the scenario. Therefore, even though scenarios might look different, they are similar with respect to some aspects. For example, in a crossing situation, not all visible participants are actually of interest for the EGO. As shown in the experiments, such rather simple, but expert-guided augmentations yield better performance.

\subsubsection{Connectivity-Based Augmentation}
\begin{figure}
    \vspace{2mm}
	\centering
	\includegraphics[width=0.45\textwidth]{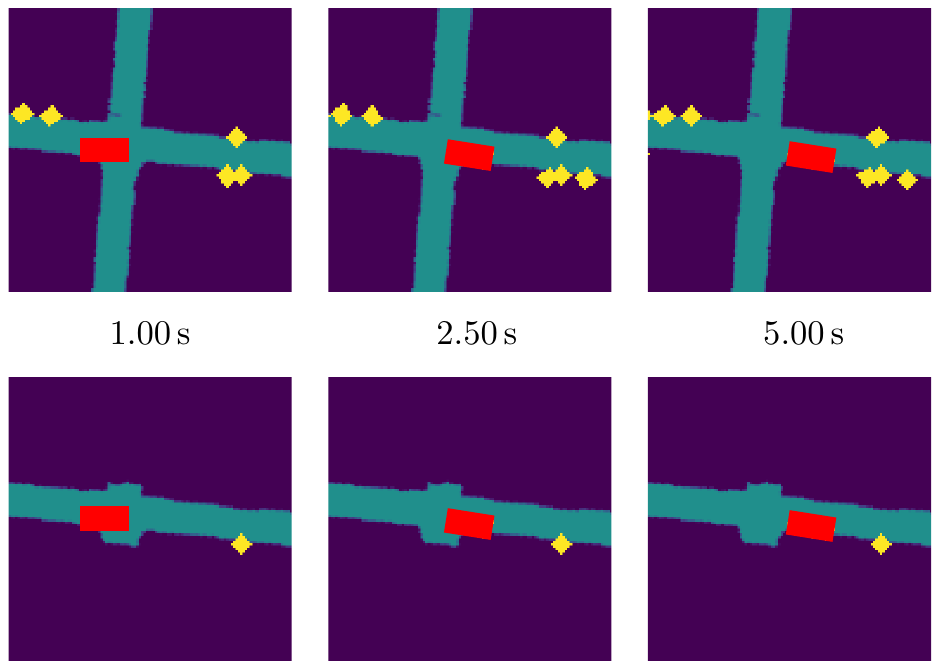}
	\caption{Connectivity-based augmentation $\mathcal{A}_\mathrm{con}$: (top) original scenario, (bottom) scenario with selected information from $\mathcal{M}$ and $\mathcal{O}$}
	\label{fig:con}
    \vspace{-5mm}
\end{figure}
For the connectivity-based augmentation, map information as well as object participants information, are dropped. The idea behind this is to retain all elements that are relevant to the EGO or to any object connected to the EGO. For this purpose, the map topology of the scenarios is utilised.

The selection of the map elements and objects to be included in the augmented scenario $S_\mathrm{con}=\mathcal{A}_\mathrm{con}(S)$ can be summarised as follows. All objects which are either directly connected to the EGO or connected to the EGO through a chain of objects are included in the augmented object list $\mathcal{O}_\mathrm{con}$. Connection here means: if they pass the same map element, a neighbouring map element or the same intersection. The map elements used for the augmentation $\mathcal{M}_\mathrm{con}$ are all elements, which are passed by the objects in $\mathcal{O}_\mathrm{con}$. And all elements connected to $\mathcal{M}_\mathrm{con}$, given the previous definition. This way, it is aimed to collect most of the relevant objects and  parts of the infrastructure, which can influence the behaviour or driving decisions.

To create the augmented scenario $S_\mathrm{con}=(\mathcal{O}_\mathrm{con},\mathcal{M}_\mathrm{con})$ as described above, the following steps are used. First, the EGO is added to the temporary object set $\mathcal{O}_\mathrm{temp}$, then all lane elements which are passed by the EGO are gathered in $\mathcal{M}_\mathrm{init}$. The set $\mathcal{M}_\mathrm{conn}$ contains all neighbouring lanes and lanes which are part of the same intersections as the lane pieces in $\mathcal{M}_\mathrm{init}$. The lane sets $\mathcal{M}_\mathrm{init}$ and $\mathcal{M}_\mathrm{conn}$ are merged to the temporary lane set $\mathcal{M}_\mathrm{temp}$. After this first run, the process is repeated until the sets $\mathcal{M}_\mathrm{temp}$ and $\mathcal{O}_\mathrm{temp}$ do not change. Therefore, the object set contains all the objects which are topologically connected to the EGO. This approach aims to consider possible interacting traffic participants. The complete procedure is also summarised in Algorithm \ref{alg:con}. An example of a resulting augmented occupancy grid sequence can be seen in Fig. \ref{fig:con}.

Let $\mathtt{inpolygon}(\mathcal{O},m)$ be a function returning true if any of the objects' trajectory points in $\mathcal{O}$ lie within the polygon $\mathcal{P}$ of $m$. Hence, if any of the objects are passing the polygon, the function returns true. The function 
$\mathtt{connected}(\mathcal{M},m)$ returns true if the element $m$ is either a neighbour to any element in $\mathcal{M}$ or the element is part of the same intersection as any element in $\mathcal{M}$.
\begin{algorithm}
    \caption{Connectivity-based augmentation $\mathcal{A}_\mathrm{con}$}\label{alg:con}
     \textbf{Input}: $S$\\
     \textbf{Output}: $S_\mathrm{con}$
    \begin{algorithmic}
    \State $\mathcal{O}_\mathrm{temp} = \lbrace o_\mathrm{EGO} \rbrace$
    \While{$\mathcal{O}_\mathrm{temp}$ or $\mathcal{M}_\mathrm{temp}$ changes}
        \State $\mathcal{M}_\mathrm{pass} = \lbrace m \in \mathcal{M}\vert \mathtt{inpolygon}(\mathcal{O}_\mathrm{temp},m) \rbrace$
        \State $\mathcal{M}_\mathrm{conn} = \lbrace m \in \mathcal{M}\vert \mathtt{connected}(\mathcal{M}_\mathrm{pass},m) \rbrace$
        \State $\mathcal{M}_\mathrm{temp} = \mathcal{M}_\mathrm{pass} \cup \mathcal{M}_\mathrm{conn}$
        \State $\mathcal{O}_\mathrm{temp} = \lbrace o \in \mathcal{O}\vert \mathtt{inpolygon}(o,m) m \in \mathcal{M}_\mathrm{temp}\rbrace$
    \EndWhile
    \State $S_\mathrm{con}=(\mathcal{O}_\mathrm{temp},\mathcal{M}_\mathrm{temp})$
    \end{algorithmic}
    \end{algorithm}

\subsubsection{Sensor-Based Augmentation}
\begin{figure}[ht]
	\centering
	\includegraphics[width=0.45\textwidth]{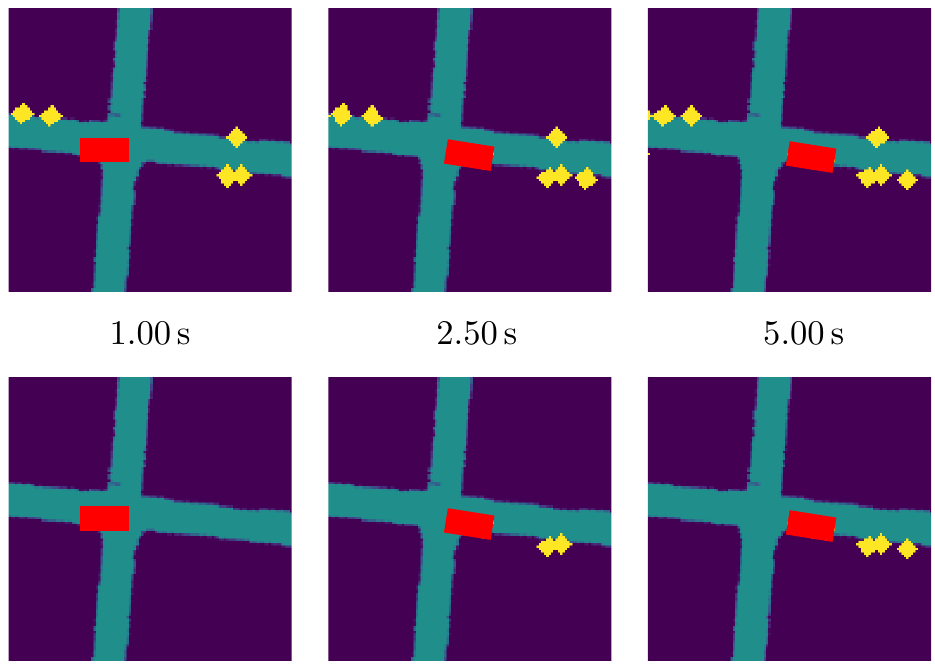}
	\caption{Sensor-based augmentation $\mathcal{A}_\mathrm{VR}$: (top) original scenario, (bottom) scenario with sensor parameters restricted to FoV as \ang{30} and range as \SI{25}{\meter}}
	\label{fig:sensr}
\end{figure}

Dependening on the sensor used in a car, the VR, formed by the FoV and range, varies. Hence, the same scenario can be perceived differently. From~\cite{Chen2021}, it can be seen that not all the traffic participants around the EGO vehicle do influence the motion planning performance of the EGO vehicle. An ideal sensor with varying VR is selected as second augmentation type.

For the sensor-based augmentation $S_\mathrm{VR}=\mathcal{A}_\mathrm{VR}(S)$, an ideal sensor fixed at the centre of the EGO vehicle is assumed. The sensor is parametrised with: maximum $\alpha_\text{max}$ and minimum $\alpha_\text{min}$ FoV, maximum $d_\text{max}$ and minimum $d_\text{min}$ range. To create a distorted view of a scenario, a random FoV, $\alpha_\text{VR}$ and range, $d_\text{VR}$ is sampled uniformly from ($\alpha_\text{min}$, $\alpha_\text{max}$) and ($d_\text{min}$,$d_\text{max}$). Traffic participants within the VR are kept in the occupancy grids and all other traffic participants are removed. An exemplary sensor based augmentation can be seen in Fig.~\ref{fig:sensr}.

If an object is inside the VR can be formulated as
\begin{equation}
    \texttt{inVR}(o(t)) = \left\lbrace \begin{array}{ll}
        1 & \text{if} -\alpha_\text{VR}<\alpha_o<\alpha_\text{VR} \wedge d_o <d_\text{VR}\\
        0 & \text{else}
    \end{array}\right.,
\end{equation}
where $\alpha_o$ is the angle from the object $o$ to the EGO's line of sight at time $t$ and $d_o(t)$ is the distance between the object $o$ and the EGO at time $t$. Given this definition, generating the augmented scenario $S_\mathrm{VR}$ can be defined as in Algorithm~\ref{alg:cap}. Hence, every timeframe is filtered for objects which are inside the current VR of the EGO.
\begin{algorithm}
	\caption{Sensor-based augmentation $\mathcal{A}_\mathrm{VR}$}\label{alg:cap}
	\textbf{Input}: $S$, $\alpha_\text{VR}=U(\alpha_\text{min},\alpha_\text{max})$, $d_\text{VR}=U(d_\text{min},d_\text{max})$\\
	\textbf{Output}: $S_\mathrm{VR}$
	\begin{algorithmic}
		\State $\mathcal{O}_\mathrm{temp} = \lbrace o_\mathrm{EGO} \rbrace$
		\For{$o$ in $\mathcal{O}$}
		\State $o_\mathrm{temp} = \lbrace\rbrace$
		\For{$t$ in $\mathrm{timestamps}$}
		\If{\texttt{inVR}($o(t)$)}
		\State $o_\mathrm{temp}\mathtt{.append}(o(t))$
		\EndIf
		\EndFor
		\State $\mathcal{O}_\mathrm{temp}\mathtt{.append}(o_\mathrm{temp})$
		\EndFor
		\State $S_\mathrm{VR}=(\mathcal{O}_\mathrm{temp},\mathcal{M})$
	\end{algorithmic}
\end{algorithm}

\section{EXPERIMENTS AND RESULTS}
\label{Sec:Exp}
This section discusses the experimental setup and analysis. The representation space generated by the self-supervised learning method with and without expert-guided augmentation is analysed with respect to the following parameters: (1) \textit{Zero-shot clustering}: Is there structure in the representation space formed by self-supervised pre-training? (2) \textit{Linear evaluation}: Is the representation space formed by self-supervised pre-training linearly separable? (3) \textit{Low-shot image classification}: Are few labelled examples enough for fine-tuning the self-supervised network for classification? (4) \textit{Representation space stability}: Are there local neighbourhood relations in the representation space? (5) \textit{Ablation}: What is the influence of various augmentations?

This section is structured as follows. The dataset used for the experimental study is described first. The baselines and the self-supervised backbone used for comparing the results of the proposed method is discussed next. Finally, the results and analysis with respect to each aspect of the representation space as discussed above are presented. 

\subsection{Dataset}

For all experiments, the Argoverse~\cite{Chang2019} dataset is used. There are a total of $333$K scenarios, each one of \SI{5}{\second} length and sampled at \SI{10}{\hertz}. All the scenarios have an EGO which is present in the scenario for the complete timespan. In order to test downstream tasks like zero-shot clustering and classification, labels for the traffic scenarios are required. To create such labels, data mining strategies are applied for each scenario. A feature vector $\bm{y}$ is extracted for each scenario as

\begin{equation}
	\bm{y} = \left[y_\text{inLeft},y_\text{inRight},y_\text{inStr},y_\text{lnChng},y_\text{straight} \right]^T,
\end{equation}
with

\begin{itemize}
	\item $y_\text{inLeft}$: if the EGO vehicle has taken a left turn in an intersection over the complete scenario
	\item $y_\text{inRight}$: if the EGO vehicle has taken a right turn in an intersection over the complete scenario
	\item $y_\text{inStr}$: if the EGO vehicle is going straight in an intersection over the complete scenario
	\item $y_\text{lnChng}$: if the EGO vehicle has done a lane change over the complete scenario
	\item $y_\text{straight}$: if the EGO vehicle is staying in the same lane over the complete scenario.
\end{itemize}

Using $\bm{y}$, the dataset $\mathcal{D}_l = \left\{(S_{1},\bm{y}_1),\ldots,(S_{M},\bm{y}_M)\right\}$ is formulated and used for testing downstream tasks. The number of classes in $\mathcal{D}_l$ can be calculated by $\mathtt{unique}(\bm{y}_1,$\ldots$,\bm{y}_M)$ leading to $26$ classes. 

\subsection{Implementation Details}

The sequence of occupancy grids $\bm{G}_i$ contains occupancy grids of size $120 \times 120$ pixels with each pixel corresponding to one meter. The EGO vehicle at time $t_0$ starts in the pixel position $(40,60)$. The grids are generated in an EGO-fixed manner. From the total \SI{5}{\second} of the scenario $4$ grids are uniformly sampled. Hence, dimension of $\bm{G}_i \in \mathbb{R}^{4\times 120\times 120}$. The backbone encoder $f(.)$ for the self-supervised pre-training is a 3D ResNet-18~\cite{He2016a}, which projects $f:\bm{G}_i\mapsto\bm{h}_i$, where $\bm{h}_i\in \mathbb{R}^{512}$. The projection network $g(.)$ is a MultiLayer Perceptron (MLP) which does the mapping $g:\bm{h}_i\mapsto\bm{z}_i$, where $\bm{z}_i\in \mathbb{R}^{2048}$. The networks are trained with both Barlow Twins and VICReg objectives with the augmentation configurations as discussed in the following section. For training, Adam optimiser is used with $lr=1e-4$ and a batch size of $200$ for Barlow Twins and a batch size of $480$ for VICReg is used for all experiments. All the experiments are carried out on a computer with Intel-i9 processor and $4\times$ RTX$6000$ GPUs.

\subsection{Baselines}

To compare the representations learnt with the expert-guided augmentations the following baselines are used:
\paragraph{Random Initialisation~(Rand. Init)}
The backbone network is initialised randomly with different random seeds. 

\paragraph{Basic Augmentations~(BaseAgt)}
Typical augmentations from CV tasks are used. The grids are augmented by randomly cropping them to $80\times 80\times 4$, randomly rotating with $2$D rotations between (-\ang{10},\ang{10}), followed by adding random noise and applying Gaussian blur. The crop size is kept big, such that most of the important scenario information is maintained.

\paragraph{Expert + Base Augmentation~(ExAgt)}
The augmentations introduced in this work are combined with the standard augmentations from \textit{BaseAgt}. In one of the views, the $\mathcal{A}_\mathrm{con}$ is applied with a probability of $0.7$ and the $\mathcal{A}_\mathrm{VR}$ is applied with a probability of $0.3$. Hence, if both augmentations are used, they are merged as defined in $\mathcal{A}_\cap$. For the other view, the  probabilities of $\mathcal{A}_\mathrm{con}$ and $\mathcal{A}_\mathrm{VR}$ are swapped. The VR is randomly sampled from ($d_\text{min}=\SI{20}{\meter}$, $d_\text{max}=\SI{100}{\meter}$) and ($\alpha_\text{min}=\ang{60}$, $\alpha_\text{max}=\ang{360}$).

\paragraph{Supervised Training~(Sup)}: The labelled dataset is used to train the network $f$ in a supervised manner.
\subsection{Metrics}
The zero-shot clustering performance is measured with unsupervised clustering accuracy~\cite{Yang2010} (ACC). The linear classifier and few-shot classification are supervised learning experiments, hence, supervised classification accuracy is used. For the representation stability experiment, the measure introduced in~\cite{Wurst2021} is used. To determine the stability measure, for each data point $k$-nearest neighbours in the representation space are chosen. The average differences with respect to features like average velocity and average trajectory displacement from the datapoint to the $k$-nearest neighbours are calculated. The lower the difference is, the more similar the features within the neighbourhood, which is an indicator for continuity and stability in the representation space. 

\subsection{Experiments with Representations}
In this section, experimental results for the clustering and the classification tasks are discussed. Each experiment is repeated $5$ times with different seeds and an average of the $5$ experiments is presented. 
\subsubsection{Zero-Shot Clustering}
This experiment is used to analyse the structure in the representation space without any fine-tuning. The trained network $f(.)$ is used to extract features $\left\{\bm{h}_1,\ldots,\bm{h}_M\right\}$ from a validation dataset of size $M$. Hierarchical clustering is performed on the extracted features due to the unbalanced nature of the classes in the dataset.

\begin{table}[ht]
    \centering
    \caption{Zero-shot clustering accuracy (higher the better)}

    \begin{tabular}{|c|c|c|c|}
        \hline
        Method & \textit{Rand. Init} &  \textit{BaseAgt}  & \textit{ExAgt} \\
        \hline \hline
        Barlow Twins & $0.29014$ & $0.3833$  & $\textbf{0.4588}$ \\
        VICReg & $0.29014$ & $0.3294$  & $\textbf{0.4142}$ \\
        \hline
    \end{tabular}
    \label{tab:zeroshotclustering}
\end{table}

From the Table~\ref{tab:zeroshotclustering}, it can be seen that when using the expert augmentation along with standard CV augmentation the zero-shot clustering accuracy improves significantly. This shows that the expert-guided augmentations are aiding in learning better representations when compared to using only standard augmentations.

\subsubsection{Linear Classifier Evaluation}
The linear classifier experiment is used to evaluate how linearly separable the traffic scenarios projected by $f(.)$ are in the representation space. For analysing this, a linear MLP layer is attached to the trained $f(.)$ and fine-tuned using the complete labelled dataset. During fine-tuning the weights of the network $f(.)$ are fixed and only the linear layer is trained.

\begin{table}[ht]
    \centering
    \caption{Linear classifier evaluation (higher the better)}

    \begin{tabular}{|c|c|c|c|}
        \hline
        Method &  \textit{BaseAgt} & \textit{ExAgt} \\
        \hline \hline
        Barlow Twins & $0.4718$ & $\textbf{0.4754}$ \\
        VICReg & $0.4345$ & $\textbf{0.4426}$  \\
        \hline
    \end{tabular}
    \label{tab:linearclasfier}
\end{table}

The linear classifier evaluation from Table~\ref{tab:linearclasfier} shows that the ExAgt leads to a comparable performance like CV augmentations.

\subsubsection{Few-Shot Classification}
The objective of the few-shot classification experiment is to test the performance of the encoder for use in downstream tasks like supervised classification. Subsets of $1$\% and $10$\% of the dataset $\mathcal{D}_l$ with the same class distribution as the validation dataset of Argoverse is selected. The pre-trained self-supervised network is fine-tuned with the $1$\% and $10$\% labelled dataset and the classification performance is reported in Table~\ref{tab:fewshot}. In both $1$\% and $10$\% settings, the fine-tuned networks which are initialised with pre-trained networks using ExAgt are able to improve the few-shot classification performance. Also, the fine-tuned networks pre-trained using both BaseAgt and ExAgt are able to outperform a supervised network trained from scratch. This shows that a self-supervised pre-training is essential in cases where the labelled data is scarce.
\begin{table}[ht]
    \centering
    \caption{Few-shot classification (higher the better). Bold: best overall, underlined: best augmentation}

    \begin{tabular}{|c|c|c|c|c|}
        \hline
        Method & \% labelled & \textit{BaseAgt}  & \textit{ExAgt} & \textit{Sup} \\
        \hline \hline
        \multirow{2}*{Barlow Twins} & $1$\%  & $0.4653$ & \underline{\textbf{0.4827}}  & $0.4131$ \\
        \cline{2-5}
         						    & $10$\%  & $0.5145$ & $\underline{\textbf{0.5453}}$  & $0.4624$ \\
         \hline						    
        \multirow{2}*{VICReg} & $1$\%  & $0.3844$ & $\underline{0.3877}$ & $\textbf{0.4131}$ \\
		\cline{2-5}
		& $10$\%  & $0.4439$ & $\underline{0.4448}$  & $\textbf{0.4624}$ \\
        \hline
    \end{tabular}
    \label{tab:fewshot}
\end{table}

\subsubsection{Representation Space Stability}

The representation space stability is used to measure the local neighbourhood relations of the data in the representation space. To investigate the representation stability, for each data point, $K$ neighbours in the representation space are considered. The average difference with respect to a selected feature e.\,g., average velocity and trajectory displacement, between the selected data point to all the $K$ neighbours is calculated. The average difference for each datapoint is again averaged over the complete dataset to get the feature stability measure~$\Delta_{...}$. Here, only the Barlow Twins method is used for the experiments.

\begin{figure}
    \vspace{2mm}
	\includegraphics[width=0.95\columnwidth]{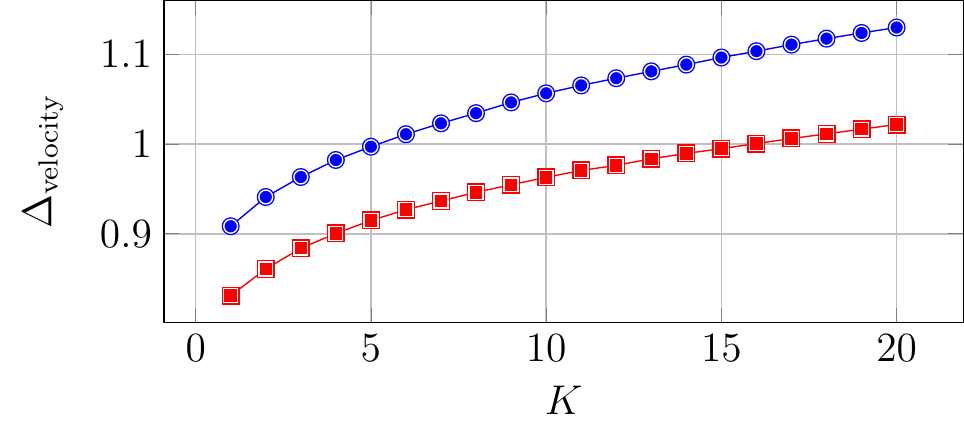}
	\includegraphics[width=0.95\columnwidth]{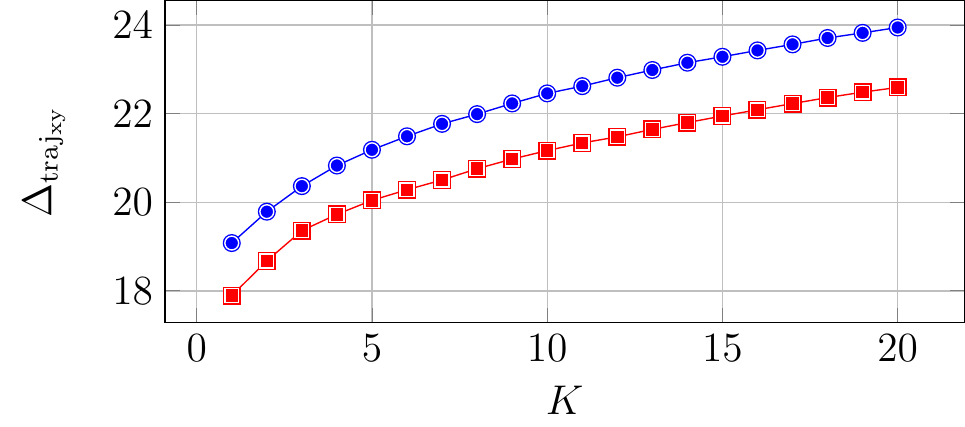}
	\includegraphics[width=0.95\columnwidth]{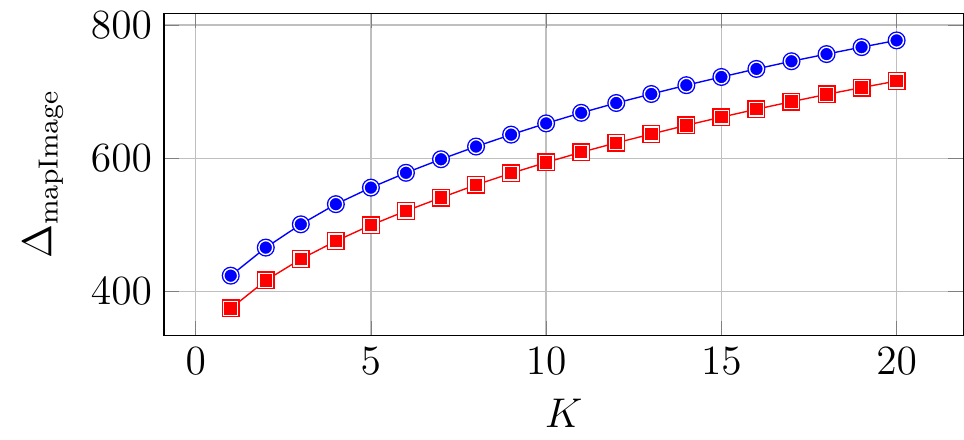}
	\caption{Representation space stability with respect to $\Delta_{\text{velocity}}$, $\Delta_{\text{traj$_\text{xy}$}}$, $\Delta_{\text{mapImage}}$ across different $K$, \textit{ExAgt}~{\expag} and \textit{BaseAgt}~{\baseag}}
	\label{fig:kfeature}
	\vspace{-5mm}
\end{figure}

In Fig.~\ref{fig:kfeature}, $\Delta_{\text{velocity}}$, the average velocity difference, $\Delta_{\text{traj$_\text{xy}$}}$, the average trajectory displacement, $\Delta_{\text{mapImage}}$, the average difference between the bird's eye view of the infrastructure image are reported across different $K$ values. With respect to every $K$ and every feature, i.\,e., average velocity, trajectory displacement, map image difference, the stability measure using ExAgt is better compared to BaseAgt. Hence, the network trained with ExAgt  has more stable local neighbourhood relations.

\subsection{Ablation Study}
\subsubsection{Augmentation Study}
This study aims to understand and analyse the effect of different augmentations for representation learning of traffic scenarios. For this the following augmentations, besides BaseAgt and ExAgt are considered:
\begin{itemize}
	\item $40$-Crop: All augmentations from BaseAgt with the random crop sized reduced to $40\times 40\times 4$,
	\item BaseAgt$+\mathcal{A}_\text{VR}$: All augmentations from BaseAgt plus the sensor-based augmentation $\mathcal{A}_\text{VR}$,
	\item BaseAgt$+\mathcal{A}_\text{con}$: All augmentations from BaseAgt plus the connectivity-based augmentation $\mathcal{A}_\text{con}$.
\end{itemize}

\begin{figure}[ht]
    \vspace{2mm}
	\centering
	\includegraphics[width=0.45\textwidth]{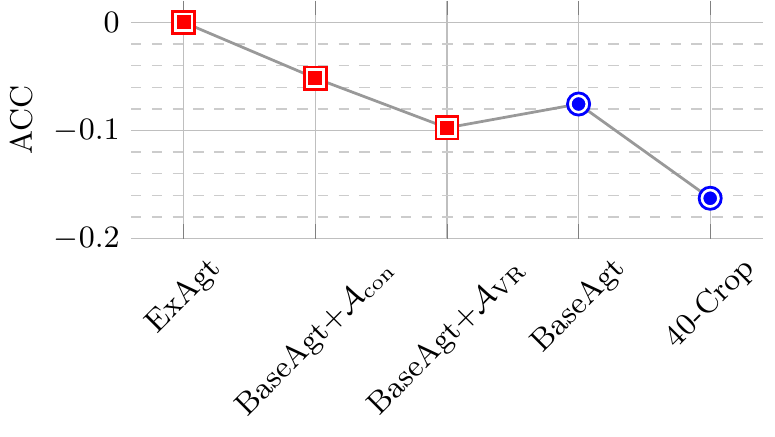}
	\caption{Reduction in unsupervised clustering accuracy from ExAgt with different augmentations, \textit{Expert-guided}~{\expag} and \textit{Basic augmentation}~{\baseag}}
	\label{fig:aug}
\end{figure}

All these experiments are compared against ExAgt with zero-shot clustering performance. The reduction of zero-shot clustering accuracy from ExAgt, when applied in the Barlow Twins method is shown in Fig.~\ref{fig:aug}. The reduction when using a reduced crop size of 40 compared to any other combination, underlines the intuition as shown in Fig.\ref{fig:baseAugmentation}. Hence, using small crop sizes with traffic scenarios should be avoided. Another conclusion that can be drawn from  Fig.~\ref{fig:aug}, is that the connectivity-based augmentation $\mathcal{A}_\mathrm{con}$ is important for the representation learning. Adding the sensor-based augmentation $\mathcal{A}_\mathrm{VR}$ leads to an even better performance, as can be seen from the value of ExAgt.

\subsubsection{VR Parameter Study}
The sensor-based augmentation $\mathcal{A}_\mathrm{VR}$ is parametrisable. Here, the impact of various VR settings is investigated. For this, unsupervised clustering performance in Barlow Twins is analysed. The default parameters used in all the experiments are ($d_\text{min}=\SI{20}{\meter}$, $d_\text{max}=\SI{100}{\meter}$) and ($\alpha_\text{min}=\ang{60}$, $\alpha_\text{max}=\ang{360}$). In this experiments, the parameters $d_\text{max}$ and  $\alpha_\text{max}$ are varied. 
\begin{table}[ht]
	\centering
	\caption{VR parameter study (higher the better)}
	
	\begin{tabular}{|c|c|c|c|}
		\hline
		Method & Range ($d_\text{min}-d_\text{max}$) &  FoV ($\alpha_\text{min}-\alpha_\text{max}$)  & ACC \\
		\hline \hline
		ExAgt & $20-100$ & $60-360$  & $\textbf{0.4588}$ \\
		ExAgt & $20-50$ & $60-360$   & $0.4069$ \\
		ExAgt & $20-100$ & $60-120$   & $0.3948$ \\
		ExAgt & $20-50$ & $60-120$   & $0.3561$ \\
		\hline
	\end{tabular}
	\label{tab:VRStudy}
\end{table}

In Table~\ref{tab:VRStudy}, the clustering performance for the various settings are shown. The setting used as default in this work shows the best overall performance. Also, it can be seen, that both, the FoV and the range are important for the performance. Hence, randomly sampling from a large VR is favourable for representation learning of traffic scenarios. 

\section{CONCLUSIONS}
\label{Sec:Con}

In this work, ExAgt a novel approach to augment traffic scenarios with expert knowledge is presented. Augmentation is important for self-supervised learning methods. ExAgt is used in self-supervised learning methods for learning representations of traffic scenarios. The representations learnt with ExAgt are compared with the representations learnt with standard CV augmentations.

Experiments show that using ExAgt is improving the performance in most of the downstream tasks and leads to better representation space stability. Hence, expert-guided domain-specific augmentations are important for traffic scenario representation learning. 

In future work, evaluating the pre-trained encoder for tasks like outlier detection and open-set recognition can be explored.
\bibliographystyle{IEEEtran}
\bibliography{format,ref}

\end{document}